%% file: main.tex
\documentclass[letterpaper]{article} % DO NOT CHANGE THIS
\usepackage[preprint]{aaai2027}  % preprint: show authors, no AAAI copyright slug
\usepackage[hyphens]{url}  % DO NOT CHANGE THIS
\usepackage{graphicx} % DO NOT CHANGE THIS
\usepackage{natbib}  % DO NOT CHANGE THIS AND DO NOT ADD ANY OPTIONS TO IT
\usepackage{caption} % DO NOT CHANGE THIS AND DO NOT ADD ANY OPTIONS TO IT
\usepackage{booktabs}
\usepackage{colortbl}
\usepackage{amsmath}
\usepackage{amsfonts}
\definecolor{methodrow}{gray}{0.93}

\title{Learned Readout Tokens for Extracting Large Language Model Confidence}

\author{
    Nima Seifi, Martin Ester\\
}
\affiliations{
    School of Computing Science, Simon Fraser University\\
    8888 University Drive, Burnaby, BC V5A 1S6, Canada\\
    \{nseifihe, ester\}@sfu.ca
}

\newcommand{\method}{\textsc{ScratchProbe}}
\newcommand{\conf}{\texttt{<CONF>}}
\newcommand{\ptrue}{P(True)}

\begin{document}

\maketitle

\begin{abstract}
    Large language models (LLMs) are increasingly used within everyday applications where their responses have significant downstream impacts. For responsible deployment of such systems, we desire insight into how much we can trust their individual outputs. A popular approach for estimating LLM confidence is to train probes on the hidden states of the LLM's tokens to predict the correctness of the model's output. These probes typically read hidden states from tokens that are already in the model's vocabulary. In this work, we introduce a new approach to confidence probes that learn a new readout token and utilize its hidden states to train the probe. We show that our learned readout approach \method{} outperforms its existing-token counterpart on a variety of datasets and LLMs. We also extend our method to a multi-token approach which we show provides further performance improvements. Finally, since the learned readout changes only the position a probe reads from, any existing probe can adopt it. Applied to a state-of-the-art probe, it improves AUROC on every backbone, in domain and out of domain.

\end{abstract}

%% ============================================================
\section{Introduction}
%% ============================================================

As large language models (LLMs) are increasingly utilized in high-stakes applications, the real-world consequences of their outputs are becoming more significant \cite{magesh2025hallucinationfree,morganstanley2024genai,jabarian2025voiceai,qiu2025physicianllm}. We seek a \textit{calibrated} confidence score: one that approximates the probability of whether a generated answer is correct. A well-calibrated model enables systems to automate decisions when the model is confident, and defer to a higher authority (human or stronger model) when uncertainty is high. This becomes especially important in high-stakes applications where the cost of an error is high.

In traditional machine learning, the confidence of a prediction is the softmax of the model's logit. LLMs inherit this convention, but an LLM is trained to predict the next token. Its logits therefore score how likely a token string is, not how likely the answer is correct. As a result, logit-derived confidence is often poorly calibrated \cite{jiang2021calibration,lin2022teaching,plaut2025probabilities,lovering2025numeric,tian2023justask}. This motivates investigating alternative signals for confidence estimation. Hidden states carry more information than the output distribution, so prior works train a probe on hidden states of existing tokens \cite{azaria2023internal, burns2023latent, orgad2025knowmore}. However, the same next-token objective shaped these hidden states, so they encode what helps predict the next token, not what predicts answer correctness. To address this, we introduce a dedicated token whose hidden state is optimized to surface information about answer correctness to the probe.

In this work, we introduce \method{}, a confidence-probing approach that jointly trains the embedding of a dedicated \conf{} readout token and a probe to predict answer correctness. In a causal language model, every token attends to all tokens that precede it. A readout token appended after the generated answer therefore has access to the complete question and the complete answer, so we place the \conf{} token at this position (Figure~\ref{fig:overview}). The same property means a single readout token is not the only option: tokens inserted before the readout can also see the full context. We therefore extend \method{} by inserting multiple trainable \textit{scratch} tokens before the final \conf{} readout. Prior dedicated-token methods such as Self-REF and GrACE update the backbone weights to produce the confidence token \cite{chuang2024selfref,Zhang2026GRACE}, and such fine-tuning degrades performance outside the training domain \cite{biderman2024lora}. We therefore freeze the LLM parameters and optimize only the added token embedding(s) and the probe.

\paragraph{Contributions}
\begin{itemize}
    \item We propose \method{}, a novel approach to confidence-probing that jointly trains a probe with token embeddings to make confidence estimates.
    \item Across three backbones and seven datasets, we show the single learned readout token on average outperforms the equivalent last-token probe.
    \item We show that our multi-token extension provides further gains over the single-token design.
    \item We demonstrate that our method can augment existing probes. Applied to Internal Inspector (I$^2$) \cite{beigi2024i2}, it improves AUROC across every backbone, in domain and out of domain.

\end{itemize}

%% ============================================================
\section{Related Work}
%% ============================================================

\subsection{Training-Free Confidence Signals}

Confidence signals that need no training read the model's own outputs. The simplest are computed from the answer's token probabilities: maximum or length-normalized log-probability and token-level entropy \cite{DBLP:conf/iclr/MalininG21,Huang_2025}, but token likelihood measures fluency and frequency as much as factual content \cite{geng2024survey}, so a fluent but wrong answer can score as high as a correct one. A second group queries the model about its answer instead of reading the answer's probabilities. \ptrue{} re-prompts the model to judge the answer and reads the probability of an affirmative token \cite{kadavath2022know}; verbalized confidence asks for a numeric score in words \cite{lin2022teaching,tian2023justask}. Both inherit the overconfidence of the model's self-reports \cite{xiong2024uncertainty}, and the verbal form also misleads the human reader, who overestimates accuracy from fluent explanations \cite{steyvers2025gap}. Self-consistency \cite{DBLP:conf/iclr/0002WSLCNCZ23,xiong2024uncertainty} avoids the self-report that \ptrue{} and verbalized confidence depend on \cite{kadavath2022know,tian2023justask}: it samples several answers and measures their agreement, at the cost of several generations per query. We use sequence likelihood, \ptrue{}, verbalized confidence, and self-consistency as baselines. \method{} shares their post-hoc placement, since it scores an answer that has already been generated, but it reads a trained hidden state rather than a decoded statement or a sampled set.

\subsection{Probing Hidden States}

The LLM probing works \cite{azaria2023internal,burns2023latent,orgad2025knowmore,beigi2024i2} apply the diagnostic classifier of \citet{alain2016probes}, a linear model trained on intermediate activations to measure what a layer makes linearly accessible. It shows that hidden states carry correctness information that the decoded text does not. These prior works all read a position that the input already contains, usually the last answer token \cite{orgad2025knowmore,zhang2025reasoning}. Prior works provide two findings that shape \method{}. First, probe quality depends on which position and layer the hidden state is taken from \cite{orgad2025knowmore,beigi2024i2}. Second, probes trained on one task only transfer to tasks that require similar skills \cite{orgad2025knowmore}. \method{} leverages these findings by \textbf{(a.)} reading from a token whose hidden state is optimized for the task and \textbf{(b.)} training on a mix of tasks to improve generalization.

\subsection{Learned Tokens}

\begin{figure*}[t]
    \centering
    \includegraphics[width=\textwidth]{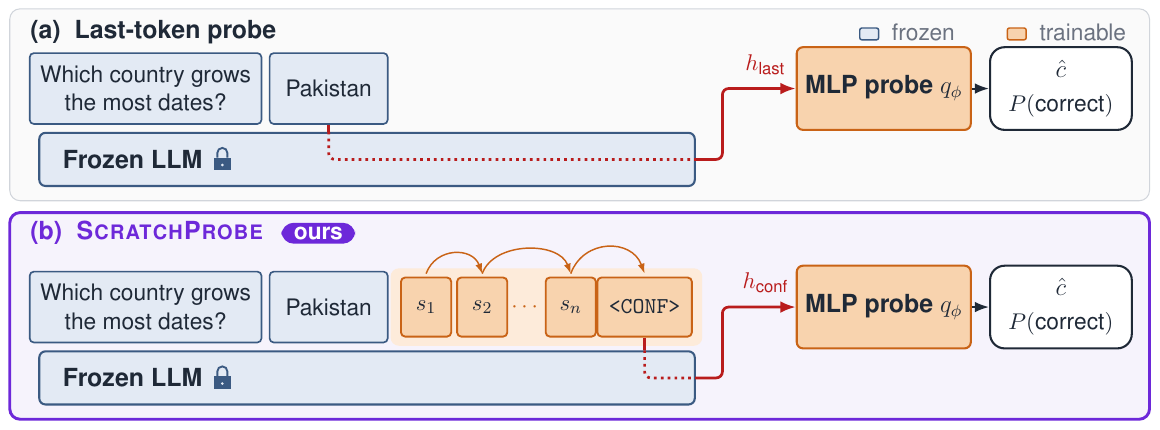}
    \caption{(a)~The standard last-token probe from prior work and (b)~\method{}. Both use the same frozen LLM and MLP probe; only the probed position differs.}
    \label{fig:overview}
\end{figure*}

Prompt tuning \cite{lester2021prompttuning} trains a token embedding against a frozen backbone, the same constraint the probes above operate under \cite{azaria2023internal,burns2023latent}. The tuned tokens are conventionally a prefix placed before the input to steer generation toward a task. \method{} uses the same mechanism with a different placement and target. A prefix precedes the answer and cannot attend to it, so \method{} places its tuned tokens after the generated answer, as a suffix (Figure~\ref{fig:overview}). The suffix is trained to supply features to a probe, not to change the output distribution, so the generated text is unchanged.

Dedicated confidence tokens have also been trained as outputs. Self-REF and the \texttt{[IDK]} token supervise the model to generate an uncertainty token and update the backbone to do so \cite{chuang2024selfref,cohen2024idk}. GrACE \cite{Zhang2026GRACE} differs from Self-REF and \texttt{[IDK]} in the readout: it adds a \texttt{<CNF>} token, trains its embedding together with LoRA adapters, and reads confidence from the LM-head probability of that token at the end of the response rather than from a generated uncertainty token. These methods obtain confidence without an auxiliary model, at the cost of modifying the weights. \method{} combines one element from each of the two preceding lines. From the output-token methods it takes the dedicated confidence token. Unlike those methods, it appends the token to the response programmatically, rather than having the model generate it. From the probing methods it takes the frozen backbone and the probe that reads the token's hidden state.

Extra token positions can also carry computation. Pause tokens appended before the answer give the model additional steps before it commits \cite{goyal2024pause}, filler tokens can perform hidden computation \cite{pfau2024filler}, and vision transformers improve when register tokens absorb global information \cite{darcet2024registers}. These three results motivate the scratch tokens in \method{}: positions inserted before the readout whose hidden states are never read directly, but which give the readout token more processed features to attend over.

%% ============================================================
\section{Method}
\label{sec:method}
%% ============================================================

\subsection{Overview}
\label{sec:overview}
Our goal is to produce a confidence estimate for an LLM's answer to a specific question. This estimate should predict whether that answer is correct.

We require access to the dataset of question-answer pairs $(q_i, a_i)$ where $q_i$ is a question and $a_i$ is its ground-truth answer. Given an LLM $p_\theta$, we sample an answer $\hat{a}_i\sim p_\theta(\cdot \mid q_i)$. We then augment this question-answer sequence with a suffix of tunable tokens. We append $n \ge 0$ \textit{scratch} tokens $s_1 \ldots s_n$, followed by the \conf{} readout token (Figure~\ref{fig:overview}). Concatenating these pieces gives the input sequence
\[u_i = [q_i]\ [\hat{a}_i]\ [s_1 \ldots s_n]\ \conf{},\]
which we pass through the LLM to obtain the hidden states of the appended tokens.

To produce a confidence estimate, we use a \textit{probe}. This is a small network $f_\phi : \mathbb{R}^d \to (0,1)$ that takes a token hidden state $h_i \in \mathbb{R}^d$ as input and outputs a predicted probability of correctness $\hat{c}_i= f_\phi(h_i)$. Prior probes read $h_i$ from a position within the generated answer as discussed in the Related Work section. We instead read it from the \conf{} position ($h_{\text{conf}}$).

A grader $G$ compares each generated answer against the ground truth and produces a binary correctness label $z_i = G(q_i, a_i, \hat{a}_i) \in \{0,1\}$. Our resulting training dataset is a set of tuples in the form $(u_i, z_i)$.

A good estimate $\hat{c}_i$ must separate correct answers from incorrect ones, and its value must approximate the observed rate of correctness. We measure the first property with the area under the ROC curve (AUROC) and the second with expected calibration error (ECE).

\subsection{New Tokens}
\label{sec:new-tokens}
The scratch tokens $s_1,\ldots,s_n$ and \conf{} do not exist in the base vocabulary, so we add them to the tokenizer and to the embedding matrix. Each scratch token has its own embedding row. A randomly initialized row would be far from the existing embeddings. This would push the hidden states off the distribution the frozen LLM was trained on. To mitigate this effect while introducing variance, we initialize each new row to the mean of the existing embedding matrix plus small Gaussian noise.

The number of scratch tokens $n$ is a hyperparameter of \method. When $n{=}0$, the augmented sequence ends with a single trainable \conf{} token. When $n{>}0$, \method{} inserts $n$ tunable scratch tokens before \conf.

\subsection{Probe}
\label{sec:probe}
The probe $f_\phi$ is a small MLP with parameters $\phi = \{W_1, b_1, W_2, b_2\}$. It reads the final-layer hidden state $h_{\text{conf}}$ at the \conf{} position and produces the confidence estimate
\[
    \hat{c} = \sigma\!\left(W_2\, \mathrm{GELU}(W_1 h_{\text{conf}} + b_1) + b_2\right) \in (0,1).
\]
For a backbone with hidden size $d$, the hidden width is $d/4$.

\subsection{Training}
\label{sec:training}
The LLM parameters $\theta$ stay frozen throughout. The only trainable parameters are the $n{+}1$ new embedding rows and the probe parameters $\phi$.

The probe output $\hat{c}_i$ is a probability and the grader label $z_i$ is binary, so we train with the binary cross-entropy (BCE) loss,
\[
    \mathcal{L}(\phi, E_{\text{new}}) = -\frac{1}{N} \sum_{i=1}^{N} \left[ z_i \log \hat{c}_i + (1 - z_i) \log (1 - \hat{c}_i) \right],
\]
where $\hat{c}_i = f_\phi(h_{\text{conf}}(u_i))$ and $E_{\text{new}}$ denotes the $n{+}1$ added embedding rows.

\subsection{Inference}
\label{sec:inference}
At test time, \method{} adds a single extra forward pass to standard generation. Given a question $q$, we first generate the answer $\hat{a} \sim p_\theta(\cdot \mid q)$ exactly as the base model would. We then append the scratch tokens and \conf{} to form $u = [q]\ [\hat{a}]\ [s_1 \ldots s_n]\ \conf{}$ and run one forward pass to obtain the final-layer hidden state $h_{\text{conf}}$. The probe maps this state to the confidence estimate $\hat{c} = f_\phi(h_{\text{conf}})$. With this pipeline, the confidence estimation step costs $n{+}1$ tokens of prefill.

\section{Experiments}
\paragraph{Training Data.}
A correctness probe trained on one task transfers only to tasks that require a similar skill \cite{orgad2025knowmore}. We therefore train on a mix of three datasets that cover different skills and answer formats: short-form question answering from TriviaQA \cite{joshi2017triviaqa}, multiple-choice science exam questions from SciQ \cite{welbl2017sciq}, and mathematical reasoning problems from GSM8K \cite{cobbe2021gsm8k}.

\paragraph{Grader.}
The datasets differ in answer format, so the grader $G$ applies a format-specific rule to each. For multiple-choice questions it extracts the option letter from the answer and compares it with the ground truth. For math problems it extracts the final number and compares it with the reference value. For free-form questions it normalizes the answer and compares it with a list of accepted aliases. A correct free-form answer can be phrased in a way that no alias covers, so answers that fail the alias match are re-graded by an LLM judge (Mistral-Small-24B-Instruct) \cite{mistral2025small}.

\paragraph{Baselines.}
\method{} does not tune the backbone, so we compare only against methods that share this property. Table~\ref{tab:macro_auroc} lists the baselines in two groups. The first group needs no training and predicts confidence from the model's outputs. The second group trains a probe on the frozen backbone with the same data as \method{}.

\begin{itemize}
    \item \textbf{Sequence likelihood} \cite{DBLP:conf/iclr/MalininG21}: the length-normalized mean token log-probability of the generated answer.
    \item \textbf{\ptrue{}} \cite{kadavath2022know}: the model is re-prompted to judge its own answer; confidence is the normalized probability of the affirmative token.
    \item \textbf{Verbalized confidence} \cite{lin2022teaching,tian2023justask}: the model is asked to state its confidence as a number from 0 to 100.
    \item \textbf{Self-consistency} \cite{DBLP:conf/iclr/0002WSLCNCZ23,xiong2024uncertainty}: $k{=}10$ answers are sampled per question at temperature 1.0; confidence is the fraction that agree with the greedily decoded answer.
    \item \textbf{Semantic entropy} \cite{DBLP:conf/iclr/KuhnGF23,DBLP:journals/nature/FarquharKKG24}: the same $k{=}10$ samples are clustered by meaning with an NLI model; confidence is $\exp(-H)$ where $H$ is the entropy over clusters.
    \item \textbf{Internal Inspector (I$^2$)} \cite{beigi2024i2}: it trains a separate 8-layer Transformer encoder on the LLM's residual stream, attention output, and feed-forward output at every layer. This makes it the most expensive baseline: it stores three states per layer for every example and trains a probe with many more parameters, whereas \method{} reads one final-layer state.
    \item \textbf{Last-token probe} \cite{azaria2023internal,orgad2025knowmore}: the same probe and loss as \method{}, but it reads the final-layer hidden state at the last answer token. Comparing against this isolates the effect of the learned read position. It is therefore the central comparison of the paper.
    \item \textbf{\method{} ($n{=}0$, ours)}: a single learned \conf{} readout token and no scratch tokens.
    \item \textbf{\method{} ($n{=}8$, ours)}: eight learned scratch tokens before the \conf{} readout.
\end{itemize}

Every method scores the same greedily generated answers, so the methods differ only in how they score, not in what answer they score.

\paragraph{Backbones.}
We evaluate our method across three instruction-tuned backbones: Llama-3.2-3B-Instruct and Llama-3.1-8B-Instruct \cite{grattafiori2024llama3}, and Phi-4-mini-instruct \cite{abouelenin2025phi4mini}. As explained in the Method section, we keep these backbone parameters frozen.

\paragraph{Evaluation.}
We evaluate in two settings. In-domain (ID) evaluation uses held-out test splits of the three training datasets (TriviaQA, SciQ, and GSM8K). Out-of-domain (OOD) evaluation uses four benchmarks that were not seen during training: MMLU \cite{hendrycks2021mmlu}, TruthfulQA \cite{lin2022truthfulqa}, CommonsenseQA (CSQA) \cite{talmor2019commonsenseqa}, and MATH-500 \cite{hendrycks2021math,lightman2024letsverify}. We report the unweighted average over the ID benchmarks and over the OOD benchmarks. The full per-dataset results are in the technical supplement (Tables~\ref{tab:supp_auroc_llama3b}--\ref{tab:supp_ece_phi4mini}). The reported numbers are the mean across 3 seeds.

\section{Results}
Table~\ref{tab:macro_auroc} reports macro-average AUROC for all methods on all three backbones. With a single learned readout token ($n{=}0$), \method{} exceeds the last-token probe on the ID and OOD macro averages of all three backbones. Adding eight scratch tokens maintains or further raises the ID average on all three backbones.

Table~\ref{tab:macro_ece} reports ECE for the same methods and the same answers as Table~\ref{tab:macro_auroc}. The training-free baselines fit no parameters, so for them the two ID/OOD groupings are only two sets of benchmarks. Their higher OOD ECE therefore comes from harder benchmarks, not from a distribution shift.

\begin{table*}[t]
    \centering
    \small\setlength{\tabcolsep}{7pt}
    \begin{tabular}{lcccccc}
        \toprule
                        & \multicolumn{2}{c}{\textbf{Llama-3.2-3B}}
                        & \multicolumn{2}{c}{\textbf{Llama-3.1-8B}}
                        & \multicolumn{2}{c}{\textbf{Phi-4-mini (3.8B)}}                                                                                                     \\
        \cmidrule(lr){2-3}\cmidrule(lr){4-5}\cmidrule(lr){6-7}
        \textbf{Method} & ID                                             & OOD               & ID                & OOD               & ID                & OOD               \\
        \midrule
        Seq.\ likelihood
                        & 0.816                                          & 0.688             & 0.829             & 0.698             & 0.771             & 0.658             \\
        \ptrue{}
                        & 0.727                                          & 0.594             & 0.719             & 0.608             & 0.783             & 0.714             \\
        Verbalized conf.
                        & 0.744                                          & 0.613             & 0.714             & 0.645             & 0.553             & 0.588             \\
        Self-consistency ($k{=}10$)
                        & 0.873                                          & 0.737             & 0.873             & 0.745             & 0.835             & \textbf{0.734}    \\
        Semantic entropy ($k{=}10$)
                        & 0.840                                          & 0.723             & 0.844             & 0.730             & 0.806             & 0.718             \\
        \midrule
        Last-token probe
                        & 0.858                                          & 0.721             & 0.875             & 0.729             & 0.893             & 0.679             \\
        Internal Inspector (I$^2$)
                        & 0.872                                          & 0.723             & \underline{0.895} & 0.730             & \textbf{0.927}    & 0.714             \\
        \rowcolor{methodrow}
        \method{} ($n{=}0$, ours)
                        & \underline{0.877}                              & \textbf{0.747}    & 0.887             & \underline{0.749} & 0.906             & 0.710             \\
        \rowcolor{methodrow}
        \method{} ($n{=}8$, ours)
                        & \textbf{0.884}                                 & \underline{0.746} & \textbf{0.897}    & \textbf{0.750}    & \underline{0.915} & \underline{0.726} \\
        \bottomrule
    \end{tabular}
    \caption{Macro-average AUROC ($\uparrow$) of \method{} and baselines on three frozen instruction-tuned backbones. \textbf{Bold} marks the best and \underline{underline} marks the second-best value in each column. Per-dataset results appear in Tables~\ref{tab:supp_auroc_llama3b}--\ref{tab:supp_auroc_phi4mini} of the technical supplement.}
    \label{tab:macro_auroc}
\end{table*}

\begin{table*}[t]
    \centering
    \small\setlength{\tabcolsep}{7pt}
    \begin{tabular}{lcccccc}
        \toprule
                        & \multicolumn{2}{c}{\textbf{Llama-3.2-3B}}
                        & \multicolumn{2}{c}{\textbf{Llama-3.1-8B}}
                        & \multicolumn{2}{c}{\textbf{Phi-4-mini (3.8B)}}                                                                                                     \\
        \cmidrule(lr){2-3}\cmidrule(lr){4-5}\cmidrule(lr){6-7}
        \textbf{Method} & ID                                             & OOD               & ID                & OOD               & ID                & OOD               \\
        \midrule
        Seq.\ likelihood
                        & 0.093                                          & 0.261             & 0.073             & 0.267             & 0.140             & 0.288             \\
        \ptrue{}
                        & 0.378                                          & 0.298             & 0.221             & 0.369             & 0.149             & 0.207             \\
        Verbalized conf.
                        & 0.107                                          & 0.183             & 0.098             & 0.308             & 0.278             & 0.407             \\
        Self-consistency ($k{=}10$)
                        & 0.072                                          & 0.151             & 0.064             & 0.154             & 0.071             & 0.161             \\
        Semantic entropy ($k{=}10$)
                        & 0.123                                          & \underline{0.120} & 0.105             & \underline{0.122} & 0.124             & \underline{0.134} \\
        \midrule
        Last-token probe
                        & 0.064                                          & \textbf{0.118}    & \underline{0.043} & 0.122             & 0.054             & 0.154             \\
        Internal Inspector (I$^2$)
                        & 0.065                                          & 0.157             & 0.043             & 0.130             & \textbf{0.037}    & 0.146             \\
        \rowcolor{methodrow}
        \method{} ($n{=}0$, ours)
                        & \underline{0.060}                              & 0.130             & 0.044             & 0.130             & 0.043             & 0.157             \\
        \rowcolor{methodrow}
        \method{} ($n{=}8$, ours)
                        & \textbf{0.059}                                 & 0.123             & \textbf{0.039}    & \textbf{0.106}    & \underline{0.038} & \textbf{0.133}    \\
        \bottomrule
    \end{tabular}
    \caption{Macro-average ECE ($\downarrow$, 15 equal-mass bins).  \textbf{Bold} marks the best and \underline{underline} marks the second-best value in each column. Per-dataset results appear in Tables~\ref{tab:supp_ece_llama3b}--\ref{tab:supp_ece_phi4mini} of the technical supplement.}
    \label{tab:macro_ece}
\end{table*}

\paragraph{Trained Readout vs Last-Token Readout}
The last-token probe and \method{} share the backbone, the probe, the loss, and the training data, so a difference between them comes only from the probed position. With respect to AUROC (Table~\ref{tab:macro_auroc}), \method{} ($n{=}0$) exceeds the last-token probe in all six ID and OOD columns. With respect to ECE (Table~\ref{tab:macro_ece}), \method{} ($n{=}8$) has lower ECE in five of the six columns. These results support our central hypothesis: a learned readout position carries more information about answer correctness than the last answer token. Table~\ref{tab:confidence_example} shows one such case: for a wrong MMLU answer from Llama-3.1-8B, the last-token probe assigns 0.81 confidence and \method{} ($n{=}8$) assigns 0.17.

The comparison with I$^2$ is mixed. I$^2$ reads three states from every layer and trains an 8-layer transformer probe. Therefore, it is a much costlier probe to train. Despite this cost gap, \method{} ($n{=}8$) stays within 0.012 AUROC of I$^2$ on every ID column, exceeds it on every OOD column, and has lower ECE in five of the six columns.

\begin{table}[t]
    \centering
    \small
    \begin{tabular}{@{}lr@{}}
        \toprule
        \multicolumn{2}{@{}c@{}}{\textbf{Input-Output Example}}                  \\
        \midrule
        \multicolumn{2}{@{}p{\columnwidth}@{}}{%
            \textbf{Question:} Divided government frequently results in:

            A) reapportionment of voting districts;

            B) amendments to the Constitution;

            C) conflict between states and the federal government;

            D) delays in the confirmation of those nominated to be ambassadors.} \\ \\
        \multicolumn{2}{@{}p{\columnwidth}@{}}{%
            \textbf{Model Answer:} C \quad \textbf{Ground Truth:} D \quad
            \textbf{Grader:} Incorrect}         \\
        \midrule
        \textbf{Probe}      & \textbf{Confidence}                                \\
        Last-token          & 0.81                                               \\
        \method{} ($n{=}8$) & 0.17                                               \\
        \bottomrule
    \end{tabular}
    \caption{Input and output example. The last-token probe is confident in a wrong answer; \method{} is not.}
    \label{tab:confidence_example}
\end{table}

\begin{figure}[t]
    \centering
    \includegraphics[width=\columnwidth]{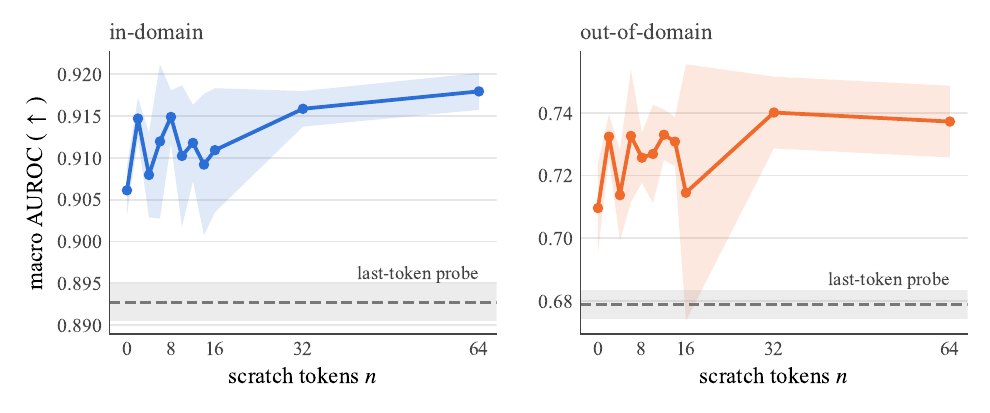}
    \caption{Macro AUROC on Phi-4-mini against $n$. The dashed line is the last-token probe.}
    \label{fig:sweep_auroc}
\end{figure}

\begin{figure}[t]
    \centering
    \includegraphics[width=\columnwidth]{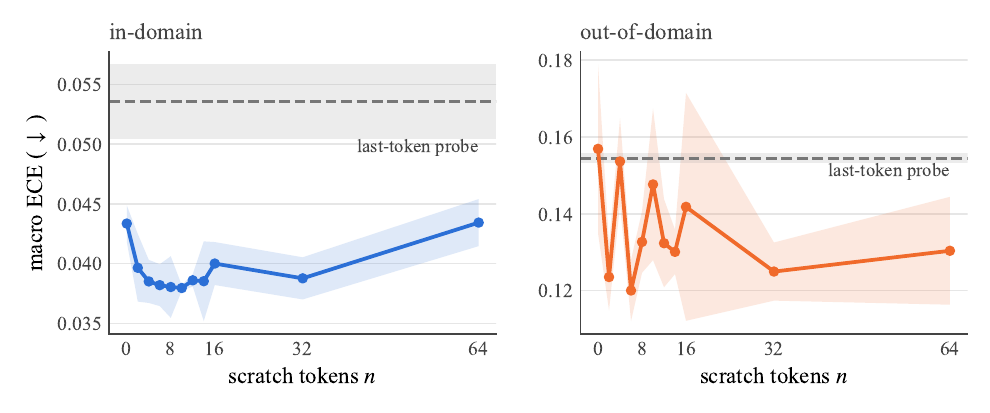}
    \caption{Macro ECE on Phi-4-mini against $n$. The dashed line is the last-token probe.}
    \label{fig:sweep_ece}
\end{figure}

\paragraph{The Impact of $n$}
Figures~\ref{fig:sweep_auroc} and~\ref{fig:sweep_ece} show the results of varying the number of scratch tokens $n$ on Phi-4-mini. This $n$-sweep shows that most of \method{}'s gain over the last-token probe is captured by $n{=}0$. In domain, $n{=}0$ recovers 60\% of the total AUROC improvement while the scratch tokens account for the remaining 40\%. These results are similar OOD.

We leave further investigation of the role and utility of these scratch tokens to future work, and we offer one direction. Probe-based methods, including \method{}, score a single generated answer and never see the alternative answers that consistency-based methods sample. The scratch tokens are a natural place to encode this signal. One could distill consistency-based information into them by training each scratch token to encode one sampled answer, so that the \conf{} readout receives consistency-based information in a single forward pass.

\begin{table*}[t]
    \centering
    \small\setlength{\tabcolsep}{4pt}
    \begin{tabular}{lcccccccccccc}
        \toprule
                       & \multicolumn{6}{c}{\textbf{AUROC} ($\uparrow$)}
                       & \multicolumn{6}{c}{\textbf{ECE} ($\downarrow$)}                                                                                                                       \\
        \cmidrule(lr){2-7}\cmidrule(lr){8-13}
                       & \multicolumn{2}{c}{Llama-3.2-3B}
                       & \multicolumn{2}{c}{Llama-3.1-8B}
                       & \multicolumn{2}{c}{Phi-4-mini}
                       & \multicolumn{2}{c}{Llama-3.2-3B}
                       & \multicolumn{2}{c}{Llama-3.1-8B}
                       & \multicolumn{2}{c}{Phi-4-mini}                                                                                                                                        \\
        \cmidrule(lr){2-3}\cmidrule(lr){4-5}\cmidrule(lr){6-7}
        \cmidrule(lr){8-9}\cmidrule(lr){10-11}\cmidrule(lr){12-13}
        \textbf{Probe} & ID                                              & OOD            & ID             & OOD            & ID             & OOD            & ID & OOD & ID & OOD & ID & OOD \\
        \midrule
        I$^2$ (last answer token)
                       & 0.872                                           & 0.723          & 0.895          & 0.730          & 0.927          & 0.714
                       & 0.065                                           & 0.157          & 0.043          & 0.130          & \textbf{0.037} & \textbf{0.146}                                  \\
        \rowcolor{methodrow}
        I$^2$ + \method{} tokens ($n{=}8$)
                       & \textbf{0.900}                                  & \textbf{0.754} & \textbf{0.906} & \textbf{0.742} & \textbf{0.935} & \textbf{0.757}
                       & \textbf{0.055}                                  & \textbf{0.137} & \textbf{0.039} & \textbf{0.112} & 0.045          & 0.173                                           \\
        \bottomrule
    \end{tabular}
    \caption{Macro-average AUROC ($\uparrow$) and ECE ($\downarrow$) of I$^2$ with its original last-token readout compared against its \method{} variant.}
    \label{tab:i2_tokens}
\end{table*}

\begin{figure}[t]
    \centering
    \includegraphics[width=\columnwidth]{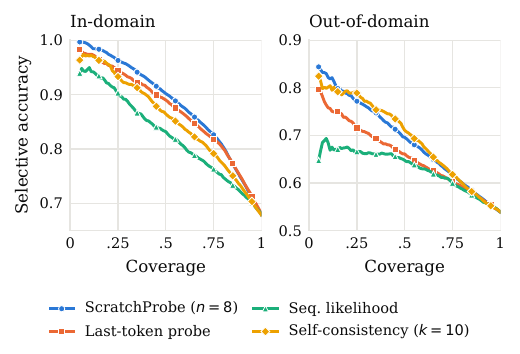}
    \caption{Selective accuracy against coverage on Phi-4-mini. At each coverage threshold, the most confident fraction of answers is kept and the accuracy of the kept set is reported.}
    \label{fig:risk_coverage}
\end{figure}

\paragraph{Risk-Coverage Tradeoff}
A confidence score can be used to decide which answers to present and which to withhold. Figure~\ref{fig:risk_coverage} visualizes how that would look for Phi-4-mini. In domain, withholding the 20\% least-confident answers raises accuracy on the remaining set from 0.68 to 0.80, and withholding half raises it to 0.90. In production, we can therefore choose the coverage we must satisfy and estimate the accuracy of our resulting system.

\paragraph{Learned Suffix Tokens Transfer to Other Probes} \method{} changes only the position a probe reads from. In principle, any probe can therefore read from a learned suffix instead of the last answer token. I$^2$ is the strongest probe-based baseline in Table~\ref{tab:macro_auroc}, so we test this claim on it. We build a \method{} variant of I$^2$ that differs from the original only in its readout: the probe reads from the \conf{} token after eight scratch tokens. As in \method{}, the suffix embeddings are trained jointly with the I$^2$ probe. Table~\ref{tab:i2_tokens} reports the result, and Table~\ref{tab:supp_i2} in the technical supplement gives the per-dataset values. The variant exceeds I$^2$ in AUROC on every backbone, ID and OOD. It has lower ECE on both Llama backbones but higher ECE on Phi-4-mini. This result suggests that a learned suffix is a cheap addition to existing probe methods. Whether the gain extends to probes for tasks outside confidence estimation is not yet clear and could be a fruitful direction for future work.

\section{Limitations}

Our backbones span two model families at 3--8B scale. Whether the gain of a learned readout over the last answer token holds at larger scale, or for base models without instruction tuning, is untested. The training mix is short-form question answering with a single gradable answer. Long-form generation, partially correct answers, and tasks without a reference answer are therefore areas for future exploration. We also do not explain what the scratch tokens compute. Their contribution is measured only through the downstream metrics, so the mechanism remains a hypothesis. Finally, \method{} reads hidden states and extends the embedding matrix, so it requires white-box access and does not apply to API-only models.

\section{Conclusion}

The hidden states that existing correctness probes read are shaped by the next-token objective, and prior probes have accepted this mismatch. Our work targets the read position as a trainable component instead. A single learned readout token improved the standard last-token probe in AUROC on every backbone, and tuning eight scratch tokens before the readout token added further gains in AUROC and ECE. The same technique improved the AUROC and ECE of an existing state-of-the-art probe. Therefore, the readout position offers an independent axis for probe improvement at the cost of $n{+}1$ tokens of prefill.

\section*{Acknowledgments}
Compute resources were provided by the Digital Research Alliance of Canada (Compute Canada). Our implementation uses PyTorch, Hugging Face Transformers, and Hugging Face PEFT.

\bibliography{bib}

\input{supplement}

\end{document}

%% file: supplement.tex
% Technical supplement, appended to re-write.tex after the references.
% Tables under supp_tables/ are generated by scripts/make_supplement_tables.py.
\clearpage
\appendix
\setcounter{table}{0}
\renewcommand{\thetable}{S\arabic{table}}
\section*{Technical Supplement}
\section{Per-Dataset Results}
\label{sec:supp_full}

The main paper reports unweighted macro averages over the three in-domain (ID) and four out-of-domain (OOD) benchmarks. Tables~\ref{tab:supp_auroc_llama3b}--\ref{tab:supp_auroc_phi4mini} give the AUROC of every method on every benchmark, one table per backbone. Tables~\ref{tab:supp_ece_llama3b}--\ref{tab:supp_ece_phi4mini} give the corresponding ECE. The ID and OOD columns of each table reproduce the macro averages of the main paper.

The per-dataset view explains two patterns that the macro averages hide. First, self-consistency owes its OOD AUROC to MATH-500. It leads \method{} ($n{=}8$) on MATH-500 on every backbone and on GSM8K on both Llama backbones, and trails \method{} on TriviaQA and SciQ on every backbone by 0.03 to 0.12. Second, the lower OOD AUROC of every probe on Phi-4-mini comes from MMLU and MATH-500, where the last-token probe falls to 0.605 and 0.742. \method{} recovers most of that gap on both benchmarks.

All probe rows use the same frozen backbone, no LoRA, and the BCE loss of the main paper with no class balancing. All methods score the same greedily generated answers. The sequence-likelihood ECE is computed on the raw length-normalized likelihood, as in the main paper.

\input{supp_tables/auroc_llama3b}
\input{supp_tables/auroc_llama8b}
\input{supp_tables/auroc_phi4mini}
\input{supp_tables/ece_llama3b}
\input{supp_tables/ece_llama8b}
\input{supp_tables/ece_phi4mini}

\section{Per-Dataset Results for I$^2$ with Learned Suffix Tokens}
\label{sec:supp_i2}

Table~\ref{tab:supp_i2} expands the I$^2$ comparison of the main paper to every benchmark. The learned suffix raises I$^2$'s AUROC on all 21 backbone--benchmark pairs. On ECE the variant is lower or equal on 11 of the 14 Llama pairs and higher on all seven Phi-4-mini pairs, with the largest Phi-4-mini increases on MMLU and CSQA.

The \method{} variant of I$^2$ was trained once, with seed 42, so its rows carry no standard deviation. The I$^2$ baseline rows report three encoder seeds.

\input{supp_tables/i2_suffix}

\section{Baseline Implementation Details}
\label{sec:supp_baselines}

All baselines score the greedily decoded answer that the frozen backbone produced for the evaluation prompt. The prompts below are passed through each backbone's chat template as a system message followed by a user message.

\paragraph{Sequence likelihood.}
The score is $\exp$ of the mean token log-probability over the answer tokens under the frozen backbone. ECE is reported on this raw score.

\paragraph{\ptrue{}.}
The backbone is re-prompted with the system message \emph{``You evaluate whether proposed answers to questions are correct.''} and the user message
\begin{quote}\small
Question: \{question\}\\
Proposed answer: \{answer\}\\
Is the proposed answer correct? Reply with exactly one letter: A if it is correct, B if it is incorrect.
\end{quote}
The confidence is the softmax over the logits of the tokens ``A'' and ``B'' at the first generated position, taken at ``A''.

\paragraph{Verbalized confidence.}
The system message is \emph{``You assess your own answers. Reply with only a number between 0 and 100.''} and the user message is
\begin{quote}\small
Question: \{question\}\\
Your answer: \{answer\}\\
How confident are you that your answer is correct? Reply with only a number between 0 and 100.
\end{quote}
The reply is decoded greedily for at most 8 tokens. The first number in the reply is divided by 100. A reply with no number is assigned 0.5.

\paragraph{Self-consistency and semantic entropy.}
Both methods draw $k{=}10$ additional answers per question at temperature 1.0 with top-$p$ 1.0, using the same prompt as the greedy answer. The samples are clustered by bidirectional entailment with DeBERTa-v2-xlarge-MNLI. For self-consistency, the confidence is the fraction of the ten samples that fall in the greedy answer's cluster. For semantic entropy, the confidence is $\exp(-H)$, where $H$ is the entropy of the cluster distribution. Both methods were run once per backbone.

\paragraph{Internal Inspector (I$^2$).}
No official implementation is available, so we re-implement the method. At the readout position, the residual stream, the attention output, and the feed-forward output of every layer are stacked into an $L{\times}3{\times}d$ tensor. An 8-layer Transformer encoder with a learned layer-position embedding and mean pooling, followed by a 3-layer MLP head, is trained with supervised-contrastive loss plus cross-entropy (temperature $0.1$). Training uses AdamW with learning rate $10^{-3}$, weight decay $10^{-4}$, batch size 64, up to 15 epochs with early stopping on validation cross-entropy. For the \method{} variant, the $n{+}1$ suffix embedding rows are trained jointly with the encoder at the same learning rate.

\paragraph{Grader and LLM judge.}
Multiple-choice answers are graded by option letter. Math answers are graded by comparing the extracted final number with the reference. Free-form answers are normalized and matched against the accepted aliases. Answers that fail the alias match are sent to Mistral-Small-24B-Instruct with the system message \emph{``You grade short answers. Given a question, one or more reference answers, and a candidate answer, decide whether the candidate is correct. Minor wording differences do not matter; the meaning must match a reference answer. Reply with exactly one word: CORRECT or INCORRECT.''} The user message lists the question, up to ten reference answers, and the candidate answer. The same judge and prompt label the training data and the evaluation data.

%% file: supp_tables/auroc_llama3b.tex
\begin{table*}[t]
    \centering
    \footnotesize\setlength{\tabcolsep}{1.4pt}
    \begin{tabular}{lccccccccc}
        \toprule
                        & \multicolumn{3}{c}{\textbf{In-domain}} & \multicolumn{4}{c}{\textbf{Out-of-domain}} & \multicolumn{2}{c}{\textbf{Macro avg.}} \\
        \cmidrule(lr){2-4}\cmidrule(lr){5-8}\cmidrule(lr){9-10}
        \textbf{Method} & TriviaQA & SciQ & GSM8K & MMLU & TruthfulQA & CSQA & MATH-500 & ID & OOD \\
        \midrule
        Seq.\ likelihood & 0.839 & 0.893 & 0.716 & \textbf{0.750} & \textbf{0.704} & \textbf{0.759} & 0.537 & 0.816 & 0.688 \\
        \ptrue{} & 0.800 & 0.670 & 0.712 & 0.537 & 0.575 & 0.586 & 0.677 & 0.727 & 0.594 \\
        Verbalized conf. & 0.799 & 0.760 & 0.672 & 0.604 & 0.581 & 0.563 & 0.704 & 0.744 & 0.613 \\
        Self-consistency ($k{=}10$) & 0.862 & 0.796 & \textbf{0.962} & 0.722 & 0.606 & 0.719 & \textbf{0.900} & 0.873 & 0.737 \\
        Semantic entropy ($k{=}10$) & 0.860 & 0.794 & \underline{0.867} & 0.717 & 0.632 & 0.718 & 0.825 & 0.840 & 0.723 \\
        \midrule
        Last-token probe & 0.863\,{\tiny$\pm$0.001} & 0.872\,{\tiny$\pm$0.002} & 0.838\,{\tiny$\pm$0.003} & 0.723\,{\tiny$\pm$0.000} & 0.615\,{\tiny$\pm$0.009} & 0.717\,{\tiny$\pm$0.003} & 0.831\,{\tiny$\pm$0.016} & 0.858\,{\tiny$\pm$0.001} & 0.721\,{\tiny$\pm$0.005} \\
        Internal Inspector (I$^2$) & \textbf{0.893}\,{\tiny$\pm$0.002} & 0.886\,{\tiny$\pm$0.011} & 0.837\,{\tiny$\pm$0.006} & 0.737\,{\tiny$\pm$0.008} & 0.603\,{\tiny$\pm$0.013} & 0.725\,{\tiny$\pm$0.016} & 0.828\,{\tiny$\pm$0.020} & 0.872\,{\tiny$\pm$0.002} & 0.723\,{\tiny$\pm$0.003} \\
        \rowcolor{methodrow}
        \method{} ($n{=}0$, ours) & 0.887\,{\tiny$\pm$0.005} & \underline{0.893}\,{\tiny$\pm$0.005} & 0.853\,{\tiny$\pm$0.011} & 0.735\,{\tiny$\pm$0.020} & \underline{0.665}\,{\tiny$\pm$0.022} & 0.740\,{\tiny$\pm$0.013} & 0.850\,{\tiny$\pm$0.011} & \underline{0.877}\,{\tiny$\pm$0.006} & \textbf{0.747}\,{\tiny$\pm$0.010} \\
        \rowcolor{methodrow}
        \method{} ($n{=}8$, ours) & \underline{0.891}\,{\tiny$\pm$0.001} & \textbf{0.905}\,{\tiny$\pm$0.006} & 0.855\,{\tiny$\pm$0.013} & \underline{0.737}\,{\tiny$\pm$0.003} & 0.641\,{\tiny$\pm$0.036} & \underline{0.744}\,{\tiny$\pm$0.017} & \underline{0.863}\,{\tiny$\pm$0.019} & \textbf{0.884}\,{\tiny$\pm$0.006} & \underline{0.746}\,{\tiny$\pm$0.006} \\
        \bottomrule
    \end{tabular}
    \caption{AUROC ($\uparrow$) on Llama-3.2-3B-Instruct, per dataset. Probe rows report mean\,$\pm$\,std over three training seeds. Sequence likelihood, \ptrue{}, and verbalized confidence are averaged over the same three seeds and have negligible spread. Self-consistency and semantic entropy use $k{=}10$ samples per question in a single run. Bold marks the best and underline the second-best value in each column.}
    \label{tab:supp_auroc_llama3b}
\end{table*}

%% file: supp_tables/auroc_llama8b.tex
\begin{table*}[t]
    \centering
    \footnotesize\setlength{\tabcolsep}{1.4pt}
    \begin{tabular}{lccccccccc}
        \toprule
                        & \multicolumn{3}{c}{\textbf{In-domain}} & \multicolumn{4}{c}{\textbf{Out-of-domain}} & \multicolumn{2}{c}{\textbf{Macro avg.}} \\
        \cmidrule(lr){2-4}\cmidrule(lr){5-8}\cmidrule(lr){9-10}
        \textbf{Method} & TriviaQA & SciQ & GSM8K & MMLU & TruthfulQA & CSQA & MATH-500 & ID & OOD \\
        \midrule
        Seq.\ likelihood & 0.862 & 0.903 & 0.721 & \textbf{0.784} & \textbf{0.660} & \textbf{0.760} & 0.588 & 0.829 & 0.698 \\
        \ptrue{} & 0.854 & 0.569 & 0.734 & 0.560 & 0.525 & 0.569 & 0.778 & 0.719 & 0.608 \\
        Verbalized conf. & 0.788 & 0.762 & 0.591 & 0.651 & 0.561 & 0.648 & 0.717 & 0.714 & 0.645 \\
        Self-consistency ($k{=}10$) & 0.881 & 0.800 & \textbf{0.940} & 0.766 & 0.621 & 0.695 & \textbf{0.899} & 0.873 & 0.745 \\
        Semantic entropy ($k{=}10$) & 0.878 & 0.799 & 0.855 & 0.760 & \underline{0.643} & 0.694 & 0.824 & 0.844 & 0.730 \\
        \midrule
        Last-token probe & 0.894\,{\tiny$\pm$0.002} & 0.902\,{\tiny$\pm$0.001} & 0.828\,{\tiny$\pm$0.011} & 0.757\,{\tiny$\pm$0.001} & 0.570\,{\tiny$\pm$0.009} & 0.732\,{\tiny$\pm$0.002} & 0.856\,{\tiny$\pm$0.008} & 0.875\,{\tiny$\pm$0.004} & 0.729\,{\tiny$\pm$0.005} \\
        Internal Inspector (I$^2$) & \underline{0.913}\,{\tiny$\pm$0.000} & \textbf{0.919}\,{\tiny$\pm$0.008} & 0.853\,{\tiny$\pm$0.007} & 0.766\,{\tiny$\pm$0.012} & 0.549\,{\tiny$\pm$0.020} & \underline{0.756}\,{\tiny$\pm$0.020} & 0.850\,{\tiny$\pm$0.008} & \underline{0.895}\,{\tiny$\pm$0.005} & 0.730\,{\tiny$\pm$0.005} \\
        \rowcolor{methodrow}
        \method{} ($n{=}0$, ours) & 0.908\,{\tiny$\pm$0.002} & 0.896\,{\tiny$\pm$0.008} & 0.857\,{\tiny$\pm$0.004} & 0.754\,{\tiny$\pm$0.008} & 0.634\,{\tiny$\pm$0.043} & 0.730\,{\tiny$\pm$0.014} & \underline{0.875}\,{\tiny$\pm$0.009} & 0.887\,{\tiny$\pm$0.002} & \underline{0.749}\,{\tiny$\pm$0.010} \\
        \rowcolor{methodrow}
        \method{} ($n{=}8$, ours) & \textbf{0.915}\,{\tiny$\pm$0.000} & \underline{0.903}\,{\tiny$\pm$0.003} & \underline{0.872}\,{\tiny$\pm$0.008} & \underline{0.769}\,{\tiny$\pm$0.006} & 0.604\,{\tiny$\pm$0.014} & 0.755\,{\tiny$\pm$0.007} & 0.873\,{\tiny$\pm$0.004} & \textbf{0.897}\,{\tiny$\pm$0.001} & \textbf{0.750}\,{\tiny$\pm$0.002} \\
        \bottomrule
    \end{tabular}
    \caption{AUROC ($\uparrow$) on Llama-3.1-8B-Instruct, per dataset. Probe rows report mean\,$\pm$\,std over three training seeds. Sequence likelihood, \ptrue{}, and verbalized confidence are averaged over the same three seeds and have negligible spread. Self-consistency and semantic entropy use $k{=}10$ samples per question in a single run. Bold marks the best and underline the second-best value in each column.}
    \label{tab:supp_auroc_llama8b}
\end{table*}

%% file: supp_tables/auroc_phi4mini.tex
\begin{table*}[t]
    \centering
    \footnotesize\setlength{\tabcolsep}{1.4pt}
    \begin{tabular}{lccccccccc}
        \toprule
                        & \multicolumn{3}{c}{\textbf{In-domain}} & \multicolumn{4}{c}{\textbf{Out-of-domain}} & \multicolumn{2}{c}{\textbf{Macro avg.}} \\
        \cmidrule(lr){2-4}\cmidrule(lr){5-8}\cmidrule(lr){9-10}
        \textbf{Method} & TriviaQA & SciQ & GSM8K & MMLU & TruthfulQA & CSQA & MATH-500 & ID & OOD \\
        \midrule
        Seq.\ likelihood & 0.844 & 0.820 & 0.651 & 0.684 & \underline{0.665} & \underline{0.737} & 0.545 & 0.771 & 0.658 \\
        \ptrue{} & 0.844 & 0.891 & 0.614 & 0.687 & \textbf{0.685} & 0.682 & 0.802 & 0.783 & 0.714 \\
        Verbalized conf. & 0.566 & 0.583 & 0.510 & 0.545 & 0.644 & 0.507 & 0.657 & 0.553 & 0.588 \\
        Self-consistency ($k{=}10$) & 0.854 & 0.796 & 0.856 & \underline{0.709} & 0.602 & 0.732 & \textbf{0.893} & 0.835 & \textbf{0.734} \\
        Semantic entropy ($k{=}10$) & 0.853 & 0.791 & 0.774 & 0.706 & 0.643 & 0.727 & 0.798 & 0.806 & 0.718 \\
        \midrule
        Last-token probe & 0.867\,{\tiny$\pm$0.001} & 0.902\,{\tiny$\pm$0.001} & 0.908\,{\tiny$\pm$0.007} & 0.605\,{\tiny$\pm$0.009} & 0.645\,{\tiny$\pm$0.005} & 0.723\,{\tiny$\pm$0.001} & 0.742\,{\tiny$\pm$0.030} & 0.893\,{\tiny$\pm$0.002} & 0.679\,{\tiny$\pm$0.005} \\
        Internal Inspector (I$^2$) & \textbf{0.906}\,{\tiny$\pm$0.001} & \textbf{0.927}\,{\tiny$\pm$0.005} & \textbf{0.948}\,{\tiny$\pm$0.003} & 0.682\,{\tiny$\pm$0.023} & 0.604\,{\tiny$\pm$0.016} & \textbf{0.757}\,{\tiny$\pm$0.002} & 0.812\,{\tiny$\pm$0.027} & \textbf{0.927}\,{\tiny$\pm$0.002} & 0.714\,{\tiny$\pm$0.003} \\
        \rowcolor{methodrow}
        \method{} ($n{=}0$, ours) & 0.877\,{\tiny$\pm$0.005} & 0.900\,{\tiny$\pm$0.005} & \underline{0.941}\,{\tiny$\pm$0.002} & \textbf{0.715}\,{\tiny$\pm$0.016} & 0.633\,{\tiny$\pm$0.024} & 0.704\,{\tiny$\pm$0.009} & 0.787\,{\tiny$\pm$0.033} & 0.906\,{\tiny$\pm$0.003} & 0.710\,{\tiny$\pm$0.014} \\
        \rowcolor{methodrow}
        \method{} ($n{=}8$, ours) & \underline{0.892}\,{\tiny$\pm$0.004} & \underline{0.913}\,{\tiny$\pm$0.007} & 0.939\,{\tiny$\pm$0.003} & 0.704\,{\tiny$\pm$0.014} & 0.637\,{\tiny$\pm$0.007} & 0.722\,{\tiny$\pm$0.005} & \underline{0.840}\,{\tiny$\pm$0.024} & \underline{0.915}\,{\tiny$\pm$0.003} & \underline{0.726}\,{\tiny$\pm$0.008} \\
        \bottomrule
    \end{tabular}
    \caption{AUROC ($\uparrow$) on Phi-4-mini-instruct (3.8B), per dataset. Probe rows report mean\,$\pm$\,std over three training seeds. Sequence likelihood, \ptrue{}, and verbalized confidence are averaged over the same three seeds and have negligible spread. Self-consistency and semantic entropy use $k{=}10$ samples per question in a single run. Bold marks the best and underline the second-best value in each column.}
    \label{tab:supp_auroc_phi4mini}
\end{table*}

%% file: supp_tables/ece_llama3b.tex
\begin{table*}[t]
    \centering
    \footnotesize\setlength{\tabcolsep}{1.4pt}
    \begin{tabular}{lccccccccc}
        \toprule
                        & \multicolumn{3}{c}{\textbf{In-domain}} & \multicolumn{4}{c}{\textbf{Out-of-domain}} & \multicolumn{2}{c}{\textbf{Macro avg.}} \\
        \cmidrule(lr){2-4}\cmidrule(lr){5-8}\cmidrule(lr){9-10}
        \textbf{Method} & TriviaQA & SciQ & GSM8K & MMLU & TruthfulQA & CSQA & MATH-500 & ID & OOD \\
        \midrule
        Seq.\ likelihood & 0.147 & 0.038 & 0.095 & 0.100 & 0.365 & 0.110 & 0.468 & 0.093 & 0.261 \\
        \ptrue{} & 0.167 & 0.407 & 0.560 & 0.337 & 0.267 & 0.350 & 0.239 & 0.378 & 0.298 \\
        Verbalized conf. & 0.134 & 0.158 & \textbf{0.031} & \underline{0.082} & 0.302 & \underline{0.094} & 0.255 & 0.107 & 0.183 \\
        Self-consistency ($k{=}10$) & \textbf{0.063} & 0.043 & 0.110 & 0.125 & 0.252 & 0.138 & \textbf{0.089} & 0.072 & 0.151 \\
        Semantic entropy ($k{=}10$) & 0.118 & 0.063 & 0.188 & \textbf{0.069} & 0.189 & 0.099 & \underline{0.125} & 0.123 & \underline{0.120} \\
        \midrule
        Last-token probe & 0.108\,{\tiny$\pm$0.010} & \textbf{0.025}\,{\tiny$\pm$0.003} & \underline{0.059}\,{\tiny$\pm$0.005} & 0.103\,{\tiny$\pm$0.007} & 0.144\,{\tiny$\pm$0.001} & \textbf{0.078}\,{\tiny$\pm$0.007} & 0.146\,{\tiny$\pm$0.011} & 0.064\,{\tiny$\pm$0.003} & \textbf{0.118}\,{\tiny$\pm$0.001} \\
        Internal Inspector (I$^2$) & 0.090\,{\tiny$\pm$0.020} & \underline{0.028}\,{\tiny$\pm$0.008} & 0.076\,{\tiny$\pm$0.016} & 0.111\,{\tiny$\pm$0.030} & 0.179\,{\tiny$\pm$0.033} & 0.101\,{\tiny$\pm$0.010} & 0.235\,{\tiny$\pm$0.020} & 0.065\,{\tiny$\pm$0.008} & 0.157\,{\tiny$\pm$0.017} \\
        \rowcolor{methodrow}
        \method{} ($n{=}0$, ours) & 0.089\,{\tiny$\pm$0.007} & 0.031\,{\tiny$\pm$0.002} & 0.059\,{\tiny$\pm$0.005} & 0.129\,{\tiny$\pm$0.046} & \textbf{0.112}\,{\tiny$\pm$0.005} & 0.105\,{\tiny$\pm$0.016} & 0.173\,{\tiny$\pm$0.017} & \underline{0.060}\,{\tiny$\pm$0.004} & 0.130\,{\tiny$\pm$0.019} \\
        \rowcolor{methodrow}
        \method{} ($n{=}8$, ours) & \underline{0.086}\,{\tiny$\pm$0.005} & 0.029\,{\tiny$\pm$0.004} & 0.061\,{\tiny$\pm$0.006} & 0.102\,{\tiny$\pm$0.017} & \underline{0.124}\,{\tiny$\pm$0.018} & 0.122\,{\tiny$\pm$0.004} & 0.143\,{\tiny$\pm$0.020} & \textbf{0.059}\,{\tiny$\pm$0.003} & 0.123\,{\tiny$\pm$0.001} \\
        \bottomrule
    \end{tabular}
    \caption{ECE ($\downarrow$) on Llama-3.2-3B-Instruct, per dataset. ECE uses 15 equal-mass bins. Probe rows report mean\,$\pm$\,std over three training seeds. Sequence likelihood, \ptrue{}, and verbalized confidence are averaged over the same three seeds and have negligible spread. Self-consistency and semantic entropy use $k{=}10$ samples per question in a single run. Bold marks the best and underline the second-best value in each column.}
    \label{tab:supp_ece_llama3b}
\end{table*}

%% file: supp_tables/ece_llama8b.tex
\begin{table*}[t]
    \centering
    \footnotesize\setlength{\tabcolsep}{1.4pt}
    \begin{tabular}{lccccccccc}
        \toprule
                        & \multicolumn{3}{c}{\textbf{In-domain}} & \multicolumn{4}{c}{\textbf{Out-of-domain}} & \multicolumn{2}{c}{\textbf{Macro avg.}} \\
        \cmidrule(lr){2-4}\cmidrule(lr){5-8}\cmidrule(lr){9-10}
        \textbf{Method} & TriviaQA & SciQ & GSM8K & MMLU & TruthfulQA & CSQA & MATH-500 & ID & OOD \\
        \midrule
        Seq.\ likelihood & 0.115 & 0.042 & 0.062 & 0.116 & 0.366 & 0.140 & 0.448 & 0.073 & 0.267 \\
        \ptrue{} & 0.157 & 0.378 & 0.128 & 0.277 & 0.439 & 0.272 & 0.489 & 0.221 & 0.369 \\
        Verbalized conf. & 0.168 & 0.033 & 0.093 & 0.207 & 0.459 & 0.099 & 0.466 & 0.098 & 0.308 \\
        Self-consistency ($k{=}10$) & \textbf{0.047} & 0.053 & 0.092 & 0.137 & 0.241 & 0.161 & \textbf{0.077} & 0.064 & 0.154 \\
        Semantic entropy ($k{=}10$) & 0.118 & 0.036 & 0.162 & 0.087 & 0.173 & 0.123 & \underline{0.103} & 0.105 & \underline{0.122} \\
        \midrule
        Last-token probe & 0.062\,{\tiny$\pm$0.008} & 0.028\,{\tiny$\pm$0.005} & 0.040\,{\tiny$\pm$0.009} & 0.115\,{\tiny$\pm$0.014} & 0.178\,{\tiny$\pm$0.011} & \textbf{0.073}\,{\tiny$\pm$0.011} & 0.123\,{\tiny$\pm$0.015} & \underline{0.043}\,{\tiny$\pm$0.001} & 0.122\,{\tiny$\pm$0.012} \\
        Internal Inspector (I$^2$) & 0.064\,{\tiny$\pm$0.026} & \textbf{0.018}\,{\tiny$\pm$0.003} & 0.048\,{\tiny$\pm$0.012} & \underline{0.086}\,{\tiny$\pm$0.031} & 0.234\,{\tiny$\pm$0.025} & \underline{0.076}\,{\tiny$\pm$0.019} & 0.123\,{\tiny$\pm$0.020} & 0.043\,{\tiny$\pm$0.012} & 0.130\,{\tiny$\pm$0.016} \\
        \rowcolor{methodrow}
        \method{} ($n{=}0$, ours) & \underline{0.061}\,{\tiny$\pm$0.004} & 0.032\,{\tiny$\pm$0.004} & \underline{0.040}\,{\tiny$\pm$0.018} & 0.168\,{\tiny$\pm$0.036} & \textbf{0.145}\,{\tiny$\pm$0.026} & 0.085\,{\tiny$\pm$0.005} & 0.122\,{\tiny$\pm$0.026} & 0.044\,{\tiny$\pm$0.008} & 0.130\,{\tiny$\pm$0.008} \\
        \rowcolor{methodrow}
        \method{} ($n{=}8$, ours) & 0.063\,{\tiny$\pm$0.012} & \underline{0.024}\,{\tiny$\pm$0.002} & \textbf{0.029}\,{\tiny$\pm$0.013} & \textbf{0.086}\,{\tiny$\pm$0.009} & \underline{0.147}\,{\tiny$\pm$0.011} & 0.087\,{\tiny$\pm$0.018} & 0.107\,{\tiny$\pm$0.023} & \textbf{0.039}\,{\tiny$\pm$0.001} & \textbf{0.106}\,{\tiny$\pm$0.005} \\
        \bottomrule
    \end{tabular}
    \caption{ECE ($\downarrow$) on Llama-3.1-8B-Instruct, per dataset. ECE uses 15 equal-mass bins. Probe rows report mean\,$\pm$\,std over three training seeds. Sequence likelihood, \ptrue{}, and verbalized confidence are averaged over the same three seeds and have negligible spread. Self-consistency and semantic entropy use $k{=}10$ samples per question in a single run. Bold marks the best and underline the second-best value in each column.}
    \label{tab:supp_ece_llama8b}
\end{table*}

%% file: supp_tables/ece_phi4mini.tex
\begin{table*}[t]
    \centering
    \footnotesize\setlength{\tabcolsep}{1.4pt}
    \begin{tabular}{lccccccccc}
        \toprule
                        & \multicolumn{3}{c}{\textbf{In-domain}} & \multicolumn{4}{c}{\textbf{Out-of-domain}} & \multicolumn{2}{c}{\textbf{Macro avg.}} \\
        \cmidrule(lr){2-4}\cmidrule(lr){5-8}\cmidrule(lr){9-10}
        \textbf{Method} & TriviaQA & SciQ & GSM8K & MMLU & TruthfulQA & CSQA & MATH-500 & ID & OOD \\
        \midrule
        Seq.\ likelihood & 0.200 & 0.095 & 0.124 & 0.300 & 0.278 & 0.190 & 0.384 & 0.140 & 0.288 \\
        \ptrue{} & 0.178 & 0.047 & 0.221 & 0.157 & 0.351 & \underline{0.109} & 0.210 & 0.149 & 0.207 \\
        Verbalized conf. & 0.447 & 0.115 & 0.272 & 0.329 & 0.565 & 0.279 & 0.454 & 0.278 & 0.407 \\
        Self-consistency ($k{=}10$) & \underline{0.064} & 0.064 & 0.086 & 0.172 & 0.236 & 0.110 & 0.125 & 0.071 & 0.161 \\
        Semantic entropy ($k{=}10$) & 0.115 & 0.054 & 0.203 & 0.114 & 0.174 & \textbf{0.085} & 0.163 & 0.124 & \underline{0.134} \\
        \midrule
        Last-token probe & 0.088\,{\tiny$\pm$0.004} & \textbf{0.022}\,{\tiny$\pm$0.004} & 0.050\,{\tiny$\pm$0.005} & 0.164\,{\tiny$\pm$0.005} & \textbf{0.142}\,{\tiny$\pm$0.004} & 0.190\,{\tiny$\pm$0.015} & 0.122\,{\tiny$\pm$0.018} & 0.054\,{\tiny$\pm$0.003} & 0.154\,{\tiny$\pm$0.001} \\
        Internal Inspector (I$^2$) & \textbf{0.058}\,{\tiny$\pm$0.008} & 0.028\,{\tiny$\pm$0.008} & 0.026\,{\tiny$\pm$0.013} & 0.139\,{\tiny$\pm$0.025} & 0.187\,{\tiny$\pm$0.020} & 0.141\,{\tiny$\pm$0.069} & \underline{0.119}\,{\tiny$\pm$0.011} & \textbf{0.037}\,{\tiny$\pm$0.009} & 0.146\,{\tiny$\pm$0.024} \\
        \rowcolor{methodrow}
        \method{} ($n{=}0$, ours) & 0.084\,{\tiny$\pm$0.003} & 0.027\,{\tiny$\pm$0.009} & \textbf{0.019}\,{\tiny$\pm$0.006} & \textbf{0.097}\,{\tiny$\pm$0.010} & 0.160\,{\tiny$\pm$0.014} & 0.224\,{\tiny$\pm$0.037} & 0.147\,{\tiny$\pm$0.044} & 0.043\,{\tiny$\pm$0.002} & 0.157\,{\tiny$\pm$0.022} \\
        \rowcolor{methodrow}
        \method{} ($n{=}8$, ours) & 0.070\,{\tiny$\pm$0.004} & \underline{0.024}\,{\tiny$\pm$0.005} & \underline{0.021}\,{\tiny$\pm$0.000} & \underline{0.097}\,{\tiny$\pm$0.007} & \underline{0.159}\,{\tiny$\pm$0.008} & 0.168\,{\tiny$\pm$0.037} & \textbf{0.107}\,{\tiny$\pm$0.010} & \underline{0.038}\,{\tiny$\pm$0.003} & \textbf{0.133}\,{\tiny$\pm$0.008} \\
        \bottomrule
    \end{tabular}
    \caption{ECE ($\downarrow$) on Phi-4-mini-instruct (3.8B), per dataset. ECE uses 15 equal-mass bins. Probe rows report mean\,$\pm$\,std over three training seeds. Sequence likelihood, \ptrue{}, and verbalized confidence are averaged over the same three seeds and have negligible spread. Self-consistency and semantic entropy use $k{=}10$ samples per question in a single run. Bold marks the best and underline the second-best value in each column.}
    \label{tab:supp_ece_phi4mini}
\end{table*}

%% file: supp_tables/i2_suffix.tex
\begin{table*}[t]
    \centering
    \scriptsize\setlength{\tabcolsep}{2.2pt}
    \begin{tabular}{lccccccccc}
        \toprule
                        & \multicolumn{3}{c}{\textbf{In-domain}} & \multicolumn{4}{c}{\textbf{Out-of-domain}} & \multicolumn{2}{c}{\textbf{Macro avg.}} \\
        \cmidrule(lr){2-4}\cmidrule(lr){5-8}\cmidrule(lr){9-10}
        \textbf{Probe} & TriviaQA & SciQ & GSM8K & MMLU & TruthfulQA & CSQA & MATH-500 & ID & OOD \\
        \midrule
        \multicolumn{10}{l}{\textbf{AUROC} ($\uparrow$)} \\
        \addlinespace[2pt]
        \multicolumn{10}{l}{\emph{Llama-3.2-3B-Instruct}} \\
        \quad I$^2$ (last answer token) & 0.893\,{\tiny$\pm$0.002} & 0.886\,{\tiny$\pm$0.011} & 0.837\,{\tiny$\pm$0.006} & 0.737\,{\tiny$\pm$0.008} & 0.603\,{\tiny$\pm$0.013} & 0.725\,{\tiny$\pm$0.016} & 0.828\,{\tiny$\pm$0.020} & 0.872\,{\tiny$\pm$0.002} & 0.723\,{\tiny$\pm$0.003} \\
        \rowcolor{methodrow}
        \quad I$^2$ + \method{} tokens ($n{=}8$) & \textbf{0.902} & \textbf{0.907} & \textbf{0.891} & \textbf{0.749} & \textbf{0.627} & \textbf{0.764} & \textbf{0.877} & \textbf{0.900} & \textbf{0.754} \\
        \addlinespace[2pt]
        \multicolumn{10}{l}{\emph{Llama-3.1-8B-Instruct}} \\
        \quad I$^2$ (last answer token) & 0.913\,{\tiny$\pm$0.000} & 0.919\,{\tiny$\pm$0.008} & 0.853\,{\tiny$\pm$0.007} & 0.766\,{\tiny$\pm$0.012} & 0.549\,{\tiny$\pm$0.020} & 0.756\,{\tiny$\pm$0.020} & 0.850\,{\tiny$\pm$0.008} & 0.895\,{\tiny$\pm$0.005} & 0.730\,{\tiny$\pm$0.005} \\
        \rowcolor{methodrow}
        \quad I$^2$ + \method{} tokens ($n{=}8$) & \textbf{0.919} & \textbf{0.920} & \textbf{0.878} & \textbf{0.774} & \textbf{0.555} & \textbf{0.766} & \textbf{0.873} & \textbf{0.906} & \textbf{0.742} \\
        \addlinespace[2pt]
        \multicolumn{10}{l}{\emph{Phi-4-mini-instruct (3.8B)}} \\
        \quad I$^2$ (last answer token) & 0.906\,{\tiny$\pm$0.001} & 0.927\,{\tiny$\pm$0.005} & 0.948\,{\tiny$\pm$0.003} & 0.682\,{\tiny$\pm$0.023} & 0.604\,{\tiny$\pm$0.016} & 0.757\,{\tiny$\pm$0.002} & 0.812\,{\tiny$\pm$0.027} & 0.927\,{\tiny$\pm$0.002} & 0.714\,{\tiny$\pm$0.003} \\
        \rowcolor{methodrow}
        \quad I$^2$ + \method{} tokens ($n{=}8$) & \textbf{0.919} & \textbf{0.928} & \textbf{0.958} & \textbf{0.775} & \textbf{0.621} & \textbf{0.772} & \textbf{0.861} & \textbf{0.935} & \textbf{0.757} \\
        \midrule
        \multicolumn{10}{l}{\textbf{ECE} ($\downarrow$)} \\
        \addlinespace[2pt]
        \multicolumn{10}{l}{\emph{Llama-3.2-3B-Instruct}} \\
        \quad I$^2$ (last answer token) & 0.090\,{\tiny$\pm$0.020} & 0.028\,{\tiny$\pm$0.008} & 0.076\,{\tiny$\pm$0.016} & \textbf{0.111}\,{\tiny$\pm$0.030} & 0.179\,{\tiny$\pm$0.033} & \textbf{0.101}\,{\tiny$\pm$0.010} & 0.235\,{\tiny$\pm$0.020} & 0.065\,{\tiny$\pm$0.008} & 0.157\,{\tiny$\pm$0.017} \\
        \rowcolor{methodrow}
        \quad I$^2$ + \method{} tokens ($n{=}8$) & \textbf{0.073} & \textbf{0.028} & \textbf{0.064} & 0.115 & \textbf{0.177} & 0.120 & \textbf{0.135} & \textbf{0.055} & \textbf{0.137} \\
        \addlinespace[2pt]
        \multicolumn{10}{l}{\emph{Llama-3.1-8B-Instruct}} \\
        \quad I$^2$ (last answer token) & 0.064\,{\tiny$\pm$0.026} & \textbf{0.018}\,{\tiny$\pm$0.003} & 0.048\,{\tiny$\pm$0.012} & 0.086\,{\tiny$\pm$0.031} & 0.234\,{\tiny$\pm$0.025} & \textbf{0.076}\,{\tiny$\pm$0.019} & 0.123\,{\tiny$\pm$0.020} & 0.043\,{\tiny$\pm$0.012} & 0.130\,{\tiny$\pm$0.016} \\
        \rowcolor{methodrow}
        \quad I$^2$ + \method{} tokens ($n{=}8$) & \textbf{0.060} & 0.018 & \textbf{0.040} & \textbf{0.084} & \textbf{0.190} & 0.077 & \textbf{0.097} & \textbf{0.039} & \textbf{0.112} \\
        \addlinespace[2pt]
        \multicolumn{10}{l}{\emph{Phi-4-mini-instruct (3.8B)}} \\
        \quad I$^2$ (last answer token) & \textbf{0.058}\,{\tiny$\pm$0.008} & \textbf{0.028}\,{\tiny$\pm$0.008} & \textbf{0.026}\,{\tiny$\pm$0.013} & \textbf{0.139}\,{\tiny$\pm$0.025} & \textbf{0.187}\,{\tiny$\pm$0.020} & \textbf{0.141}\,{\tiny$\pm$0.069} & \textbf{0.119}\,{\tiny$\pm$0.011} & \textbf{0.037}\,{\tiny$\pm$0.009} & \textbf{0.146}\,{\tiny$\pm$0.024} \\
        \rowcolor{methodrow}
        \quad I$^2$ + \method{} tokens ($n{=}8$) & 0.061 & 0.046 & 0.027 & 0.169 & 0.188 & 0.210 & 0.123 & 0.045 & 0.173 \\
        \bottomrule
    \end{tabular}
    \caption{Per-dataset AUROC ($\uparrow$) and ECE ($\downarrow$) of I$^2$ with its original last-answer-token readout and of the \method{} variant that reads from \conf{} after eight jointly trained scratch tokens. The I$^2$ rows report mean\,$\pm$\,std over three encoder seeds. The \method{} variant rows are from a single seed (42). Bold marks the better of the two rows in each column.}
    \label{tab:supp_i2}
\end{table*}